\documentclass{article}

\usepackage{arxiv}

\usepackage[utf8]{inputenc} 
\usepackage[T1]{fontenc}    
\usepackage{hyperref}       
\usepackage{url}            
\usepackage{booktabs}       
\usepackage{amsfonts}       
\usepackage{nicefrac}       
\usepackage{microtype}      
\usepackage{graphicx}
\usepackage{floatrow}

\title{\emph{Oculum afficit}: Ocular affect recognition}

\author{
    Elmar H. Langholz \\
    Department of Computer Science \\
    University of Illinois \\
    Illinois, USA \\
    \texttt{elmarhl2@illinois.edu}
}

\begin{document}
\maketitle

\begin{abstract}
Recognizing human affect and emotions is a problem that has a wide range of applications within both academia and industry. Affect and emotion recognition within computer vision primarily relies on images of faces. With the prevalence of portable devices (e.g. smart phones and/or smart glasses), acquiring user facial images requires focus, time, and precision. While existing systems work great for full frontal faces, they tend to not work so well with partial faces like those of the operator of the device when under use. Due to this, we propose a methodology in which we can accurately infer the overall affect of a person by looking at the ocular region of an individual.
\end{abstract}


\section{Introduction}
\label{sec:introduction}

Using machines to recognize human affect (a concept used in psychology to describe the experience of feeling or emotion) and emotions has always been a challenging topic that amasses huge interest from both academia and industry. There are currently a variety of ways to recognize human affects and emotions such as analysis of body language, voice intonation, and more involved methods like MRI and EEG. However, a more popular and practical approach for affect and emotion recognition is to rely on computer vision by primarily looking at images of faces to analyze facial expressions. 

In a world where portable devices \footnote{Smartphones, VR, AR and/or MR glasses like Oculus Rift, HoloLens, etc...} have increasingly gained popularity, acquiring a facial image for accurate recognition usually requires full focus, the whole face, and precise timing of capturing the image (capturing a face in transition can result in only a partial facial image). Furthermore, when a subject's face is partially occluded with masks, glasses, objects, or even a beard, the accuracy of recognizing affect and emotions decreases drastically with current models.

We believe it is possible to predict the affect of a user by aiming our attention on just part of their face. More precisely, the ocular region of a subject can convey much about an individual's affect, especially through a portable device. 

We chose the ocular region consisting of eyes and eyebrows since as human beings we have learned to evolve to communicate emotions through them \cite{mindeyes2017}. Also, eyes are rarely occluded by users when using portable devices since they must normally look at the screen (close to the camera) in order to capture their image. In fact, smart glasses are used directly in front of the eyes.

\section{Related work}
\label{sec:related_work}

Research focusing on using faces to recognize emotion while leveraging Convolutional Neural Network (CNN) has become prevalent \cite{christou2019human} and can even include attention mechanisms to try to address occlusion \cite{li2019occlusion}. Leveraging these techniques using faces to determine emotion and affect on mobile devices \cite{cnnafaceaffect2018} also exist. There is also a new focus in combining emotion recognition techniques like EEG and facial landmark localization \cite{li2019facial} to solve this problem. Furthermore, generation of facial expression from a neutral expression using Identity-free conditional
Generative Adversarial Network (IF-GAN) \cite{cai2019identity} demonstrates another use following along this area of research. In all the aforementioned, affect is normally treated as a secondary artifact.

While the prior focuses on faces, using the eyes to recognize emotions through non-neural models \cite{eyeemoreco2013} is something that has been experimented with in the past. However, it is not as prevalent as facial expression recognition. As far as we are aware, there isn't any similar research for predicting the affect of a person through their ocular region using neural-based models.

\section{Data set}
\label{sec:data_set}

There exists several data sets corresponding to facial emotion recognition through face images. Some of these are available freely while others need to be requested and consent given for use. While many focus on emotion through a categorical variable, we are interested in affect which can be recorded as a set of numerical continuous variables. As a point in hand, we are interested in the location of the different facial features so that we can focus on the corresponding ocular area which is of interest for this research.

AffectNet \cite{affectnet2017}, as its name suggests, is one of the largest providers of affect and emotion labeled data for a set of face images with a size of 122 GB. It provides approximately one million labeled images which were obtained by querying three different search engines using 1250 emotion keyword in six different languages. It is not freely available and requires consent from the owners.

The data is split in two groups: manually and automatically annotated. For this research we will start by using the manually annotated data which is further split into training and validation sets. Since no test set is provided, we instead use the provided validation data set as our test set and create our own validation data set from the training set by randomly selecting $1\%$ of the entries and setting them aside. We do this to enhance reproducibility and to allow others to perform benchmarking. Table \ref{table:data_set_split} shows the initial actual sizes regarding how the data set was split.

\begin{table}[ht]
\centering
\footnotesize
\caption{Data set split}
\label{table:data_set_split}
\begin{tabular}{rccc}
\cline{2-4}
\multicolumn{1}{l}{}  & \textbf{Training} & \textbf{Validation} & \textbf{Test} \\ \cline{2-4} 
\textbf{Initial}      & 410651            & 4149                & 5500          \\
\textbf{Preprocessed} & 317521            & 3218                & 4500          \\ \cline{2-4} 
\end{tabular}
\end{table}

Affect labels for each face are provided as two separate sets of numerical continuous variables: valence and arousal. While valence corresponds to a sense of how unpleasant/negative to pleasant/positive an event is, arousal focuses on smoothing/calming to exciting/agitating. In both cases, the values have the following range: $[-1, 1]$. Face location, in the form of a bounding box, and 68-point facial landmarks corresponding to mouth, eyes, eyebrows and face contour are provided.

\section{Methods and design}
\label{sec:methods_and_design}

\subsection{Technology and tools}
\label{subsec:technology_and_tools}

The machine learning pipeline developed for this research was written primarily using the Python \cite{rossum1995python} programming language. For data preprocessing Pandas \cite{mckinney2011pandas} was used to store the data and OpenCV \cite{opencv_library} was leveraged to perform image operations. For modeling, training, and evaluation, Keras \cite{chollet2015keras} was used with a Tensorflow \cite{tensorflow2015-whitepaper} back-end along with NumPy \cite{numpy}. Visualizations were constructed using keras-vis \cite{raghakotkerasvis}. Execution was done in an a machine with an NVIDIA GeForce GTX 1080 Ti GPU having approximately $64$ GB in ram and $800$ GB of hard drive space.

\subsection{Preprocessing}
\label{subsec:preprocessing}

\begin{figure}[ht]
\centering
\footnotesize
$\vcenter{\hbox{\includegraphics[width=0.25\textwidth]{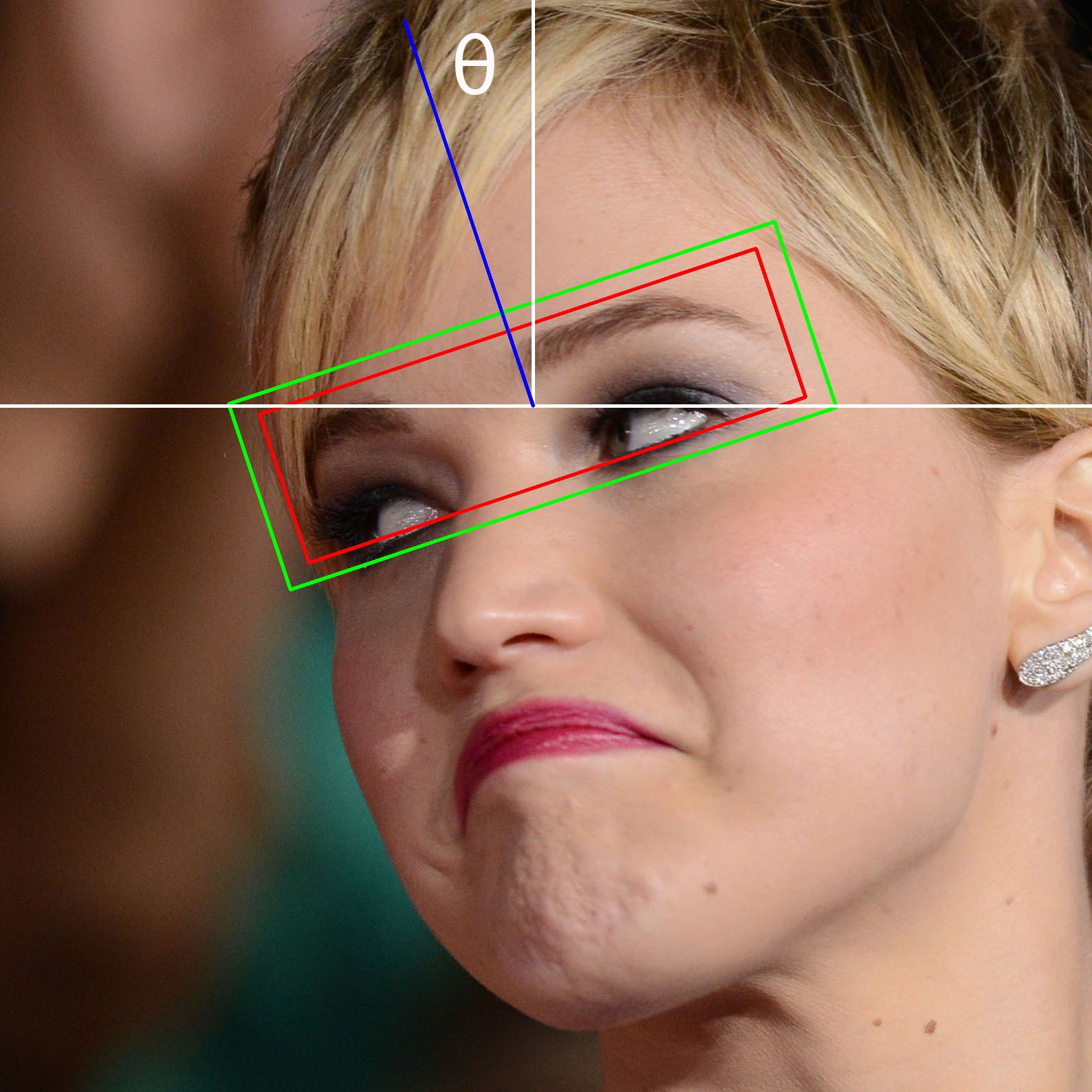}}}$
\hspace*{.2in}
$\vcenter{\hbox{\includegraphics[width=0.25\textwidth]{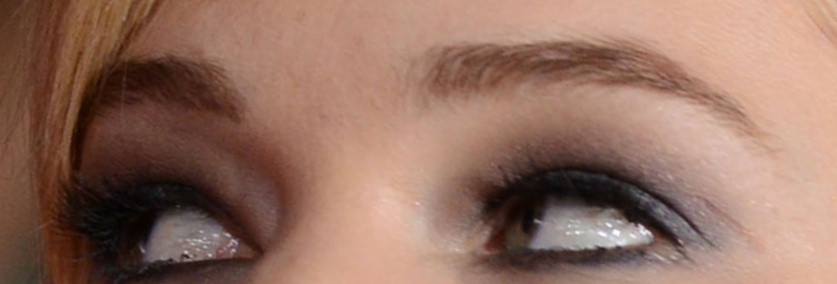}}}$
\caption{Preprocessed image example}
\label{example_image_preprocessing}
\floatfoot{The first image (left) demonstrates visually the constructed bounding box before and after expanding its size, as well as the calculated rotation $\theta$. The second image (right) is the eye slot derived from preprocessing.}
\end{figure}

We derived a new data set by extracting ocular regions from AffectNet. Using the facial landmark points (provided with the data set) corresponding to both eyes and eyebrows and the closest nose point, an initial bounding box comprised of minimum area for these points was calculated for each image. The size of bounding box was proportionally increased by $10\%$ and $25\%$, horizontally and vertically respectively, while keeping its center point the same. Since the centerline of a face rarely aligns to the horizontal of the image, most of the bounding boxes were rotated. Therefore, the degree of rotation $\theta$ was also determined and the image was rotated around the center point of the bounding box before cropping. This is depicted by the top image in Figure \ref{example_image_preprocessing}.

A rectangular ocular region image derived through the above process is called an eye slot as shown by the bottom image in Figure \ref{example_image_preprocessing}. If the corresponding arousal or valence is not within the expected documented range, then the image is skipped and not processed. If the extracted eye slot image width and height are not larger than zero or the height is larger than the width, it is also not included as part of the data set. Table \ref{table:data_set_split} shows the remaining data set size split after preprocessing.

\subsection{Augmentation and normalization}
\label{subsec:augmentation_and_normalization}

The more and varied the data we train our model with, the better it becomes with respect to accuracy and robustness. One technique used in order to increase the size of the data set as well as introduce transformations to images is data augmentation \cite{kaiming2015deepres}. While preprocessing constructs standardized eye slots offline by leveraging facial landmarks, augmentation transforms and standardizes the eye slot size at run time to construct an infinite amount of different eye slots from one.

During eye slot augmentation different types of transformations were applied to the original eye slot. For each type, a random value within a defined range was generated and used to transform the image. Table \ref{table:transformation_types_ranges_and_units} lists the different types of transformation, value ranges, and corresponding units:

\begin{table}[ht]
\centering
\footnotesize
\caption{Transformation types, value ranges and units.}
\label{table:transformation_types_ranges_and_units}
\begin{tabular}{ccc}
\hline
\textbf{Type}   & \textbf{Range}  & \textbf{Unit}     \\ \hline
Brightness      & {[0.5, 1.5]}    & Lightness (HLS)   \\
Rotation        & {[0, 5]}        & Degrees           \\
Width shift     & {[0.0, 0.10]}   & Width percentage  \\
Height shift    & {[0.0, 0.10]}   & Height percentage \\
Shear           & {[0.0, 0.01]}   & Radians           \\
Horizontal flip & -               & -                 \\ \hline
\end{tabular}
\end{table}

\begin{figure}[ht]
\centering
\includegraphics[width=0.25\textwidth]{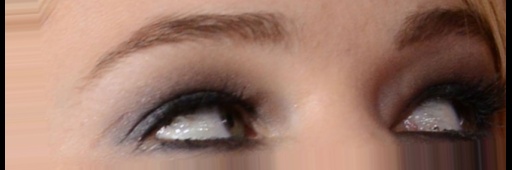}
\caption{Augmented and normalized image example}
\label{example_image_augmented_resized}
\floatfoot{An example of augmentation and normalization of the eye slot from Figure \ref{example_image_preprocessing}. A random set of transformations were applied and padding is added (in black) to the left and right to maintain the aspect ratio of the image and resize it to $512 \times 170$ pixels.}
\end{figure}

Up until this point, we retained the original size of the eye slot derived from the image. However, once the set of transformations are applied to the image, its size was normalized so that the images could be used with ease for modeling. Since we believe that each individual detail provided within the eye slot is important and relevant to predicting affect, we maintain the aspect ratio of the original eye slot and pad the image at the top and bottom and/or left and right in order to resize it to  $512 \times 170$ pixels. The default image value range of $[0, 255]$ was then normalized to a range of $[0.0, 1.0]$ with the intent to speed up calculations by using floating point operations. Figure \ref{example_image_augmented_resized} shows an example of augmentation and normalization used during model training.

\subsection{Modeling}
\label{subsec:modeling}

In computer vision, Deep Convolutional Neural Networks (DCNN) have been working extremely well on achieving state of the art results for multiple different computer vision tasks \cite{dcnnsurvey2019}, as well as face expression recognition \cite{deepfaceexpressionreco2018}. Due to this, we decided to leverage DCNN architectures to solve a dual regression problem in which our predictor variable was the eye slot images and the response variables (representing affect) were their valence and arousal. We focused on the VGGNet-like \cite{vggnet2015} and MobileNet-like \cite{mobilenet2017} architectures. The composition of these DCNN implementation configurations are shown in Table \ref{table:dcnn_configurations}. These architectures are comprised of groups called blocks.

\begin{table}[ht]
\centering
\scriptsize
\caption{DCNN configrations}
\label{table:dcnn_configurations}
\begin{tabular}{ccc}
\hline
\multicolumn{3}{|c|}{DCNN configuration}                                                                                                                                                                                                                            \\ \hline
\multicolumn{1}{|c|}{$M_{1}$}                                                       & \multicolumn{1}{c|}{$M_{2}$}                                                       & \multicolumn{1}{c|}{$M_{3}$}                                                             \\ \hline
\multicolumn{1}{|c|}{\begin{tabular}[c]{@{}c@{}}VGGNet\\ 14 layers\end{tabular}}    & \multicolumn{1}{c|}{\begin{tabular}[c]{@{}c@{}}VGGNet\\ 15 layers\end{tabular}}    & \multicolumn{1}{c|}{\begin{tabular}[c]{@{}c@{}}MobileNet\\ $\omega = 1$\end{tabular}}    \\ \hline
\multicolumn{2}{l}{}                                                                                                                                                     & \multicolumn{1}{l}{}                                                                     \\ \hline
\multicolumn{3}{|c|}{input (170 $\times$ 512 RGB image)}                                                                                                                                                                                                            \\ \hline
\multicolumn{1}{|c|}{\begin{tabular}[c]{@{}c@{}}conv3-16\\ conv3-16\end{tabular}}   & \multicolumn{1}{c|}{\begin{tabular}[c]{@{}c@{}}conv3-64\\ conv3-64\end{tabular}}   & \multicolumn{1}{c|}{conv3-32-s2}                                                         \\ \hline
\multicolumn{2}{|c|}{max pool}                                                                                                                                           & \multicolumn{1}{c|}{\begin{tabular}[c]{@{}c@{}}dw-conv3-s1\\ conv1-64-s1\end{tabular}}   \\ \hline
\multicolumn{1}{|c|}{\begin{tabular}[c]{@{}c@{}}conv3-32\\ conv3-32\end{tabular}}   & \multicolumn{1}{c|}{\begin{tabular}[c]{@{}c@{}}conv3-128\\ conv3-128\end{tabular}} & \multicolumn{1}{c|}{\begin{tabular}[c]{@{}c@{}}dw-conv3-s2\\ conv1-128-s1\end{tabular}}  \\ \hline
\multicolumn{2}{|c|}{max pool}                                                                                                                                           & \multicolumn{1}{c|}{\begin{tabular}[c]{@{}c@{}}dw-conv3-s1\\ conv1-128-s1\end{tabular}}  \\ \hline
\multicolumn{1}{|c|}{\begin{tabular}[c]{@{}c@{}}conv3-64\\ conv3-64\end{tabular}}   & \multicolumn{1}{c|}{\begin{tabular}[c]{@{}c@{}}conv3-256\\ conv3-256\end{tabular}} & \multicolumn{1}{c|}{\begin{tabular}[c]{@{}c@{}}dw-conv3-s2\\ conv1-256-s1\end{tabular}}  \\ \hline
\multicolumn{2}{|c|}{max pool}                                                                                                                                           & \multicolumn{1}{c|}{\begin{tabular}[c]{@{}c@{}}dw-conv3-s1\\ conv1-256-s1\end{tabular}}  \\ \hline
\multicolumn{1}{|c|}{\begin{tabular}[c]{@{}c@{}}conv3-128\\ conv3-128\end{tabular}} & \multicolumn{1}{c|}{\begin{tabular}[c]{@{}c@{}}conv3-512\\ conv3-512\end{tabular}} & \multicolumn{1}{c|}{\begin{tabular}[c]{@{}c@{}}dw-conv3-s2\\ conv1-512-s1\end{tabular}}  \\ \hline
\multicolumn{2}{|c|}{max pool}                                                                                                                                           & \multicolumn{1}{c|}{\begin{tabular}[c]{@{}c@{}}dw-conv3-s1\\ conv1-512-s1\end{tabular}}  \\ \hline
\multicolumn{1}{|c|}{\begin{tabular}[c]{@{}c@{}}conv3-256\\ conv3-256\end{tabular}} & \multicolumn{1}{c|}{\begin{tabular}[c]{@{}c@{}}conv3-512\\ conv3-512\end{tabular}} & \multicolumn{1}{c|}{\begin{tabular}[c]{@{}c@{}}dw-conv3-s1\\ conv1-512-s1\end{tabular}}  \\ \hline
\multicolumn{2}{|c|}{max pool}                                                                                                                                           & \multicolumn{1}{c|}{\begin{tabular}[c]{@{}c@{}}dw-conv3-s1\\ conv1-512-s1\end{tabular}}  \\ \hline
\multicolumn{1}{|c|}{\begin{tabular}[c]{@{}c@{}}conv3-512\\ conv3-512\end{tabular}} & \multicolumn{1}{c|}{\begin{tabular}[c]{@{}c@{}}conv3-512\\ conv3-512\end{tabular}} & \multicolumn{1}{c|}{\begin{tabular}[c]{@{}c@{}}dw-conv3-s1\\ conv1-512-s1\end{tabular}}  \\ \hline
\multicolumn{2}{|c|}{max pool}                                                                                                                                           & \multicolumn{1}{c|}{\begin{tabular}[c]{@{}c@{}}dw-conv3-s1\\ conv1-512-s1\end{tabular}}  \\ \hline
\multicolumn{1}{|c|}{FC-6144}                                                       & \multicolumn{1}{c|}{FC-6144}                                                       & \multicolumn{1}{c|}{\begin{tabular}[c]{@{}c@{}}dw-conv3-s2\\ conv1-1024-s1\end{tabular}} \\ \hline
\multicolumn{1}{|c|}{FC-6144}                                                       & \multicolumn{1}{c|}{FC-6144}                                                       & \multicolumn{1}{c|}{\begin{tabular}[c]{@{}c@{}}dw-conv3-s1\\ conv1-1024-s1\end{tabular}} \\ \hline
\multicolumn{1}{l|}{}                                                               & \multicolumn{1}{c|}{FC-2000}                                                       & \multicolumn{1}{l|}{global avg pool}                                                     \\ \cline{2-3} 
\end{tabular}
\end{table}

For VGGNet, a block consists of two convolution layers. Each convolution used a kernel size of $3 \times 3$. In a block, a convolution layer was followed by a batch normalization layer and a rectified linear unit (ReLU) activation layer. Following a block, a max pool of $2 \times 2$ was used. After the blocks, fully connected (or dense) layers of different sizes were added.

MobileNet blocks consist of a depth-wise convolution with a kernel size of $3 \times 3$ with either a stride of $1 \times 1$ or $2 \times 2$ followed by batch normalization and a rectified linear unit (ReLU) activation layer. To complete the block, a convolution layer consisting of a kernel size and stride of $1 \times 1$, followed by batch normalization and a rectified linear unit (ReLU) activation layer were also present. Once the blocks were completed a global average pooling layer was added

At the end of each DCNN configuration, a fully connected layer consisting of two units with a linear activation was added. One for each matching response affect variable. We do this with the intent of reusing the convolutional filters into a single model instead of having two separate models.

\subsection{Training}
\label{subsec:training}

\begin{figure}[ht]
\centering
\footnotesize
\includegraphics[width=0.5\linewidth]{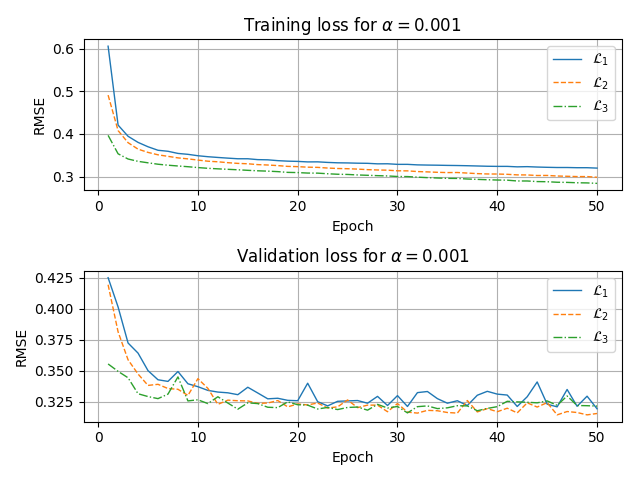}
\caption{Model loss plots}
\label{models_loss_plots}
\floatfoot{The first and second image (from top to bottom) show the loss for the training and validation corresponding to each model. The loss of $M_{1}$ is depicted by $\mathcal{L}_{1}$, $M_{2}$ loss is depicted by $\mathcal{L}_{2}$ and $M_{3}$ is $\mathcal{L}_{3}$. All models ran for $\eta = 50$ epochs.}
\end{figure}

Being a dual regression problem (we are trying to learn both valence and arousal using the same filters in one model), the mean squared error loss function was selected to optimize using Adam \cite{adamopt2015}. We varied the optimizer per run to use a learning rates $\alpha = 10^{-3}$, $\beta_{1} = 0.9$, $\beta_{2} = 0.999$, and $\epsilon = 10^{-8}$. For each run, training was performed with a batch size $\gamma = 16$. All models were run for $\eta = 50$ epochs. Their corresponding training times can be found in Table \ref{table:training_times} and their loss plots are shown in Figure \ref{models_loss_plots}.

\begin{table}[ht]
\centering
\footnotesize
\caption{Model training times}
\label{table:training_times}
\begin{tabular}{cc}
\toprule
     Model & Time (days) \\
     \midrule
     $M_{1}$ & 2.88 \\
     $M_{2}$ & 3.66 \\
     $M_{3}$ & \textbf{2.85} \\
\bottomrule
\end{tabular}
\end{table}

The model training times capture the fact that $M_{3}$ took the least amount of time to train. The difference between $M_{1}$ and $M_{3}$ is approximately 51 minutes while the difference between $M_{2}$ and $M_{3}$ is approximately 20 hours. This is highly related to the number of weights that each model has since $M_{3}$ is optimized to reduce its amount of parameters and this property can be further exploited by reducing the network width $\omega$ which is a hyper-parameter of MobileNet.

Through simple visual inspection we can observe that $M_{2}$ outperforms $M_{1}$ by a very small margin during training and validation. Between $M_{2}$ and $M_{3}$ this distinction is not that clear since during training $M_{3}$ seems to have a smaller loss than $M_{2}$. Nonetheless when looking at the validation loss plot we can see that while $M_{3}$ initially reduces is loss faster, but after a few epochs $M_{2}$ catches up while crossing each other every couple of epochs.

\section{Evaluation and results}
\label{sec:evaluation_and_results}

To evaluate the models, we calculated the root mean squared error (RMSE), Pearson's correlation coefficient (CORR), concordance correlation coefficient (CCC) \cite{lawrence1989concordance} and sign agreement metric (SAGR) \cite{nicolaou2011continuous} on the test data set. These are standard metrics and are used in other research and therefore we provide them to allow comparisons. Summarized results for valence (V) and arousal (A) can be found in Table \ref{table:models_valence_arousal_evaluation}.

\begin{table}[ht]
\centering
\footnotesize
\caption{Models valence and arousal evaluation}
\label{table:models_valence_arousal_evaluation}
\begin{tabular}{ccccccccc}
\toprule
                              & \multicolumn{2}{c}{\textbf{RMSE}}              & \multicolumn{2}{c}{\textbf{CORR}}              & \multicolumn{2}{c}{\textbf{CCC}}               & \multicolumn{2}{c}{\textbf{SAGR}}              \\
                              \cmidrule(r){2-3}
                              \cmidrule(r){4-5}
                              \cmidrule(r){6-7}
                              \cmidrule(r){8-9}
                              
                              & \multicolumn{1}{c}{V} & \multicolumn{1}{c}{A} & \multicolumn{1}{c}{V} & \multicolumn{1}{c}{A} & \multicolumn{1}{c}{V} & \multicolumn{1}{c}{A} & \multicolumn{1}{c}{V} & \multicolumn{1}{c}{A} \\ \midrule
\multicolumn{1}{c|}{$M_{1}$} & .456                   & .405                   & .513                   & .496                   & .434                   & .354                   & .672                   & .750                   \\
\multicolumn{1}{c|}{$M_{2}$} & \textbf{.444}                   & \textbf{.389}                   & \textbf{.533}                   & \textbf{.514}                   & .445                   & .404                   & \textbf{.690}                   & \textbf{.751}                   \\
\multicolumn{1}{c|}{$M_{3}$} & .466                   & .394                   & .507                   & .495                   & \textbf{.467}                   & \textbf{.425}                   & .677                   & .729                   \\
\bottomrule
\end{tabular}
\end{table}

Model $M_{2}$ outperforms the remaining models in RMSE, CORR and SAGR. However, $M_3$ outperforms the others on CCC. Due to these results we can say that $M_{2}$ will provide better generalization at the cost of more weights. Focusing on the differences between valence and arousal, $M_{2}$ performs better on arousal than valence with respect to RMSE and SAGR. However, valence is better than arousal on CORR and CCC.

In order to further understand how the model predicts these values, we decided to look at the attention of the model using the regression dense layer. This is possible by assessing the gradients of the model with respect to an input image to determine which areas of the image cause the inferred valence and arousal values to change. These areas can then be visualized through a heatmap of the input image allowing one to better understand which of these influence the model to make its inference.

\begin{figure}[ht]
\centering
\footnotesize
\includegraphics[width=0.60\textwidth]{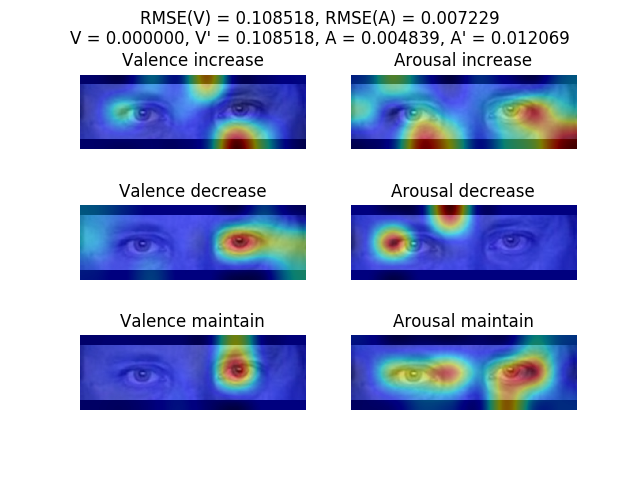}
\includegraphics[width=0.40\textwidth]{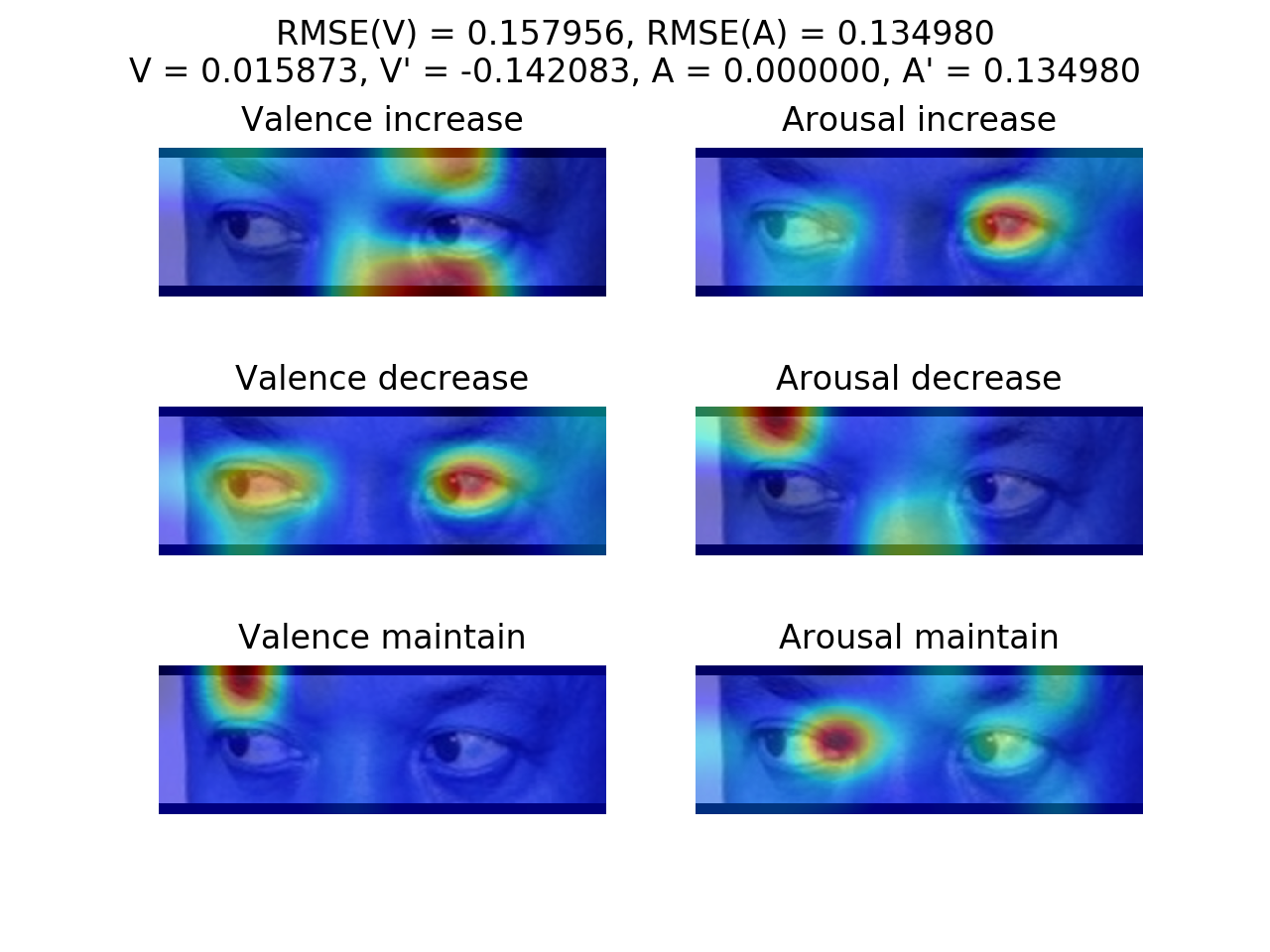}
\includegraphics[width=0.40\textwidth]{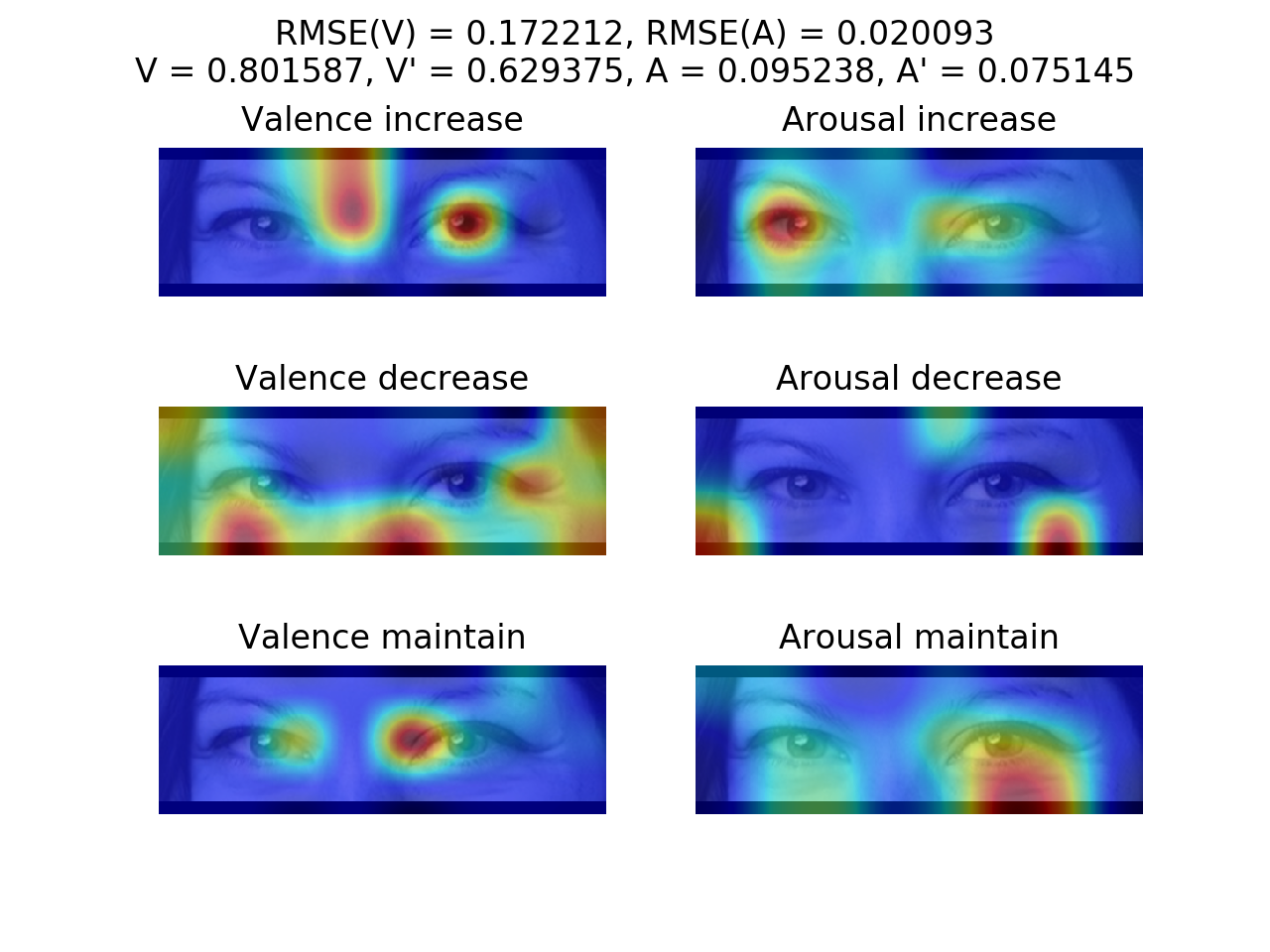}
\caption{Attention map visualization of eye slots}
\label{attention_visualization_eye_slots}
\floatfoot{The first image corresponds to a neutral valence and arousal ($V = 0.0$ and $A = 0.0$). The second and third (left to right) are two more examples of attention visualization for eye slots with different valence and arousal.}
\end{figure}

The first image of Figure \ref{attention_visualization_eye_slots} corresponds to a neutral eye slot showing the increase, decrease and maintaining valence and arousal. With it we can observe that from a neutral state the model is focusing on specific regions where change in it is likely to affect the valence or arousal. Looking closely, we see that the areas covered are varied from eyebrows to underneath the eyes. Through it, we can see that the model is trying to make decisions as how a human specialist likely would. This is reminiscent of the facial action coding system \cite{hjortsjo1969man} \cite{friesen1978facial} used to determine facial emotions \cite{ekman1982rationale}.

\section{Conclusion}
\label{sec:conclusion}

In conclusion, using this methodology we have been able to successfully produce comparable results to other affect models by focusing on the ocular region. We show that the models infer based on different areas of the ocular region. While not as exact as as facial affect models, it gives a close approximation and insight into what is possible. Nevertheless, we believe that more data and further refinement in modeling can provide even better results and more insight into this area of research.

While currently out of scope, in order to have a fully working end-to-end system, we would need to build an eye slot detector as well. Due to having the facial landmarks available, it would make sense to include in place of preprocessing so that we are able to quickly and accurately retrieve eye slots on a live system. We can use the existing data to be able to train it.

There are many interesting areas of research that this opens up such as privacy preserving ocular affect image generation or de-identification. In the future we believe that this will be important in order for agents to convey emotion while preserving ones anonymity.

\section{Acknowledgments}
\label{sec:acknowledgments}

We thank Aaron Blythe for assisting with the generation and storage of the preprocessed images due to space constraints in our setup. We also appreciate the suggested edits, questions and feedback from Zisheng Liao and Daniel Barker.

\bibliographystyle{unsrt} 
\bibliography{oculum_afficit}

\end{document}